\pdfoutput=1

\documentclass[11pt]{article}

\usepackage[]{acl}

\usepackage{times}
\usepackage{latexsym}

\usepackage[T1]{fontenc}

\usepackage[utf8]{inputenc}

\usepackage{microtype}

\usepackage{graphicx}
\usepackage{xcolor, colortbl}
\usepackage{bbm}
\usepackage{hyperref}
\usepackage{subfig}
\usepackage{amsmath}
\usepackage{amsfonts}

\definecolor{lightgreen}{HTML}{d1f8bc}
\definecolor{yellow}{HTML}{fff2cc}
\definecolor{red}{HTML}{fd6864}
\definecolor{ENC}{HTML}{fdc4c2}
\definecolor{ESC}{HTML}{c1e0f4}
\definecolor{DEC}{HTML}{ffe2ca}
\definecolor{FRC}{HTML}{acc1d9}
\definecolor{ENC_}{HTML}{fff0f0}
\definecolor{ESC_}{HTML}{eef6fc}
\definecolor{DEC_}{HTML}{fff7ef}
\definecolor{FRC_}{HTML}{e6ecf4}
\definecolor{BACK}{HTML}{f0ece9}

\title{Are Pretrained Multilingual Models Equally Fair Across Languages?}

\author{Laura Cabello Piqueras \\
  University of Copenhagen \\
  \texttt{lcp@di.ku.dk} \\\And
  Anders Søgaard \\
  University of Copenhagen \\
  \texttt{soegaard@di.ku.dk} \\}

\begin{document}
\maketitle
\begin{abstract}
Pretrained multilingual language models can help bridge the digital language divide, enabling high-quality NLP models for lower-resourced languages. Studies of multilingual models have so far focused on performance, consistency, and cross-lingual generalisation. However, with their wide-spread application in the wild and downstream societal impact, it is important to put multilingual models under the same scrutiny as monolingual models. This work investigates the group fairness of multilingual models, asking whether these models are equally fair across languages. To this end, we create a new four-way multilingual dataset of parallel cloze test examples (MozArt), equipped with demographic information (balanced with regard to gender and native tongue) about the test participants.
We evaluate three multilingual models on MozArt -- mBERT, XLM-R, and mT5 -- and show that across the four target languages, the three models exhibit different levels of group disparity, e.g., exhibiting near-equal risk for Spanish, but high levels of disparity for German.


\end{abstract}

\section{Introduction}
\label{sec:intro}
Fill-in-the-gap cloze tests \cite{Taylor1953ClozePA} ask language learners to predict what words were removed from a text and it is a ``procedure for measuring the effectiveness of communication''. 
Today, language models are trained to do the same \cite{devlin2019bert}. This has the advantage that we can now use fill-in-the-gap cloze tests to directly compare the linguistic preferences of humans and language models, e.g., to investigate task-independent sociolectal biases (group disparities) in language models \cite{zhang-etal-2021-sociolectal}. This paper presents a novel four-way parallel cloze dataset for English, French, German, and Spanish that enables apples-to-apples comparison across languages of group disparities in multilingual language models.\footnote{The language selection was given to us, because we rely on an existing word alignment dataset; see \S2.} 


\begin{table}[]
\centering
\resizebox{\columnwidth}{!}{
    \subfloat{
        \begin{tabular}{lcccc}
        & \cellcolor{ENC}\textbf{EN} & \cellcolor{ESC}\textbf{ES} & \cellcolor{DEC}\textbf{DE} & \cellcolor{FRC}\textbf{FR} \\
        {\cellcolor{BACK}\textbf{WordPiece (avg. \#tokens)}} &
          \multicolumn{1}{c}{\cellcolor{ENC_}19.7} &
          \multicolumn{1}{c}{\cellcolor{ESC_}22.0} &
          \multicolumn{1}{c}{\cellcolor{DEC_}23.6} &
          \multicolumn{1}{c}{\cellcolor{FRC_}23.1} \\
        {\cellcolor{BACK}\textbf{SentencePiece (avg. \#tokens)}} &
          \multicolumn{1}{c}{\cellcolor{ENC_}22.3} &
          \multicolumn{1}{c}{\cellcolor{ESC_}22.9} &
          \multicolumn{1}{c}{\cellcolor{DEC_}24.9} &
          \multicolumn{1}{c}{\cellcolor{FRC_}25.3} \\
        {\cellcolor{BACK}\textbf{\#Sentences}} &
          \multicolumn{1}{c}{\cellcolor{ENC_}100} &
          \multicolumn{1}{c}{\cellcolor{ESC_}100} &
          \multicolumn{1}{c}{\cellcolor{DEC_}100} &
          \multicolumn{1}{c}{\cellcolor{FRC_}100} \\
        {\cellcolor{BACK}\textbf{\#Annotations}} &
          \multicolumn{1}{c}{\cellcolor{ENC_}600} &
          \multicolumn{1}{c}{\cellcolor{ESC_}600} &
          \multicolumn{1}{c}{\cellcolor{DEC_}600} &
          \multicolumn{1}{c}{\cellcolor{FRC_}600} \\
        {\cellcolor{BACK}\textbf{\#Annotators}} &
          \multicolumn{1}{c}{\cellcolor{ENC_}60} &
          \multicolumn{1}{c}{\cellcolor{ESC_}60} &
          \multicolumn{1}{c}{\cellcolor{DEC_}60} &
          \multicolumn{1}{c}{\cellcolor{FRC_}60} \\
         \end{tabular}
    }
}
\resizebox{\columnwidth}{!}{
    \subfloat{
        \begin{tabular}{lcccc}
        \rowcolor{BACK}\textbf{Demographics} &
        \multicolumn{4}{l}{ 
            \begin{tabular}[c]{@{}l@{}}
            id$\_$u, id$\_$s, gender, age, nationality, \\ first language, fluent languages, \\ current country of residence, \\ country of birth, time taken
            \end{tabular}
        }
        \end{tabular}
    }
}
\caption{MozArt details. The average number of tokens per sentence is reported using WordPiece and SentecePiece. The bottom row lists the demographic attributes shared; id\_u refers to user id (anonymised) and id\_s to sentence id.}
\label{tab:stats}
\end{table}

Language models induced from historical data are prone to implicit biases \cite{zhao2017men, chang-etal-2019-bias, mehrabi2021art}, e.g., as a result of the over-representation of male-dominated text sources such as Wikipedia and newswire \cite{hovy-sogaard-2015-tagging}. This may lead to language models that are {\em unfair} to groups of users in the sense that they work better for some groups rather than others \cite{zhang-etal-2021-sociolectal}. Multilingual language models can be unfair to their training languages in similar ways \cite{choudhury2021how,wan2022fairness,wang2021assessing}, but this work goes beyond previous work in evaluating whether multilingual language models are {\em equally fair to demographic groups across languages}. 

To this end, we create MozArt, a multilingual dataset of fill-in-the-gap sentences covering four languages (English, French, German and Spanish). The sentences reflect diastratic variation within each language and can be used to compare biases in pretrained language models (PLMs) across languages. We study the influence of four demographic groups, i.e., the cross-product of our annotators' gender  -- male ({\em M}) or female ({\em F})\footnote{None of our annotators identified as non-binary.} -- and first language -- native ({\em N}) or non-native ({\em NN}).\footnote{See \citet{john-schmitz, ffaez} for discussion of the native/non-native speaker dichotomy. Participants were asked ``What is your first language?'' and ``Which of the following languages are you fluent in?''. We use {\em native} ({\em N}) for people whose first language coincides with the example sentences, and non-native ({\em NN}) otherwise, without any sociocultural implications.} Table~\ref{tab:stats} presents a summary of dataset characteristics.

\section{Dataset}
\label{sec:data}

We introduce MozArt, a four-way multilingual cloze test dataset with annotator demographics. We sampled 100 sentence quadruples from each of the four languages (English, French, German, Spanish) in the corpus provided for the WMT 2006 Shared Task.\footnote{\href{https://www.statmt.org/wmt06/shared-task/}{\nolinkurl{www.statmt.org/wmt06/shared-task}}} The data was extracted from the publicly available Europarl corpus \cite{koehn-2005-europarl} and enhanced with word-level bitext alignments \cite{koehn-monz-2006-manual}. 
The word alignments are important for what follows. We manually verify that sentences make sense out of context and use the data to generate {\em comparable cloze examples}, e.g.:\\[1pt]

{\small 
\begin{tabular}{ll}
     en&[MASK] that deplete the ozone layer\\
     es&[MASK] que agotan la capa de ozono\\
de&[MASK], die zum Abbau der Ozonschicht führen\\
fr&[MASK] appauvrissant la couche d'ozone
\end{tabular} 
}

\noindent We only mask words which are (i) aligned by one-to-one alignments, and which are (ii) either nouns, verbs, adjectives or adverbs.\footnote{We use spaCy's part-of-speech tagger \cite{spacy2} to predict the syntactic categories of the input words.} We mask one word in each sentence and verify that one-to-one alignments exist in all languages. Following \citet{suzanne-kleijn}, we rely on part-of-speech information to avoid masking words that are {\em too} predictable, e.g., auxiliary verbs or constituents of multi-word expressions, or words that are {\em un}-predictable, e.g., proper names and technical terms. 

Annotators were recruited using Prolific.\footnote{\url{prolific.co}} We applied eligibility criteria to balance our annotators across demographics. Participants were asked to report (on a voluntary basis) their demographic information regarding gender and languages spoken. Each eligible participant was presented with 10 cloze examples. We collected answers from 240 annotators, 60 per language batch, divided in four balanced demographic groups (gender $\times$ native language). We made sure that each sentence had at least six annotations. 
Annotation guidelines for each language were given in that language, to avoid bias and ensure a minimum of language understanding for non-native speakers. We manually filtered out spammers to ensure data quality. 

The dataset is made publicly available at \href{http://github.com/coastalcph/mozart}{\nolinkurl{github.com/coastalcph/mozart}} under a CC-BY-4.0 license. We include all the demographic attributes of our annotators as per agreement with the annotators. The full list of protected attributes is found in Table~\ref{tab:stats}. We hope MozArt will become a useful resource for the community, also for evaluating the fairness of language models across other attributes than gender and native language.

\section{Experimental Setup}
\label{sec:exp}

\paragraph{Models} We evaluate three PLMs: mBERT \cite{devlin2019bert}, XLM-RoBERTa/XLM-R \cite{conneau2020unsupervised}, and mT5 \cite{xue2021mt5}.\footnote{We use the base models available from \href{https://huggingface.co/models}{\nolinkurl{huggingface.co/models}}. We report results using uncased mBERT, since it performed better on our data than its cased sibling.} 
All three models were trained with a masked language modelling objective. mBERT differs from XLM-R and mT5 in including a next sentence prediction objective \cite{devlin2019bert}. mT5 differs from mBERT and XLM-R in allowing for consecutive spans of input tokens to be masked \cite{raffel2020exploring}. 
We adopt beam search decoding with early stopping and constrain the generation to single words. This enables better correlation of mT5's output with our group preferences. t-SNE plots are included in Appendix ~\ref{sec:appendix_tsne} to show how languages are distributed in the PLM vector spaces.

\paragraph{Metrics} We use several metrics to compare how the PLMs align with group preferences across languages. 
These include top-k precision $\mathrm{P@k}$ with k=\{1, 5\}, mean reciprocal rank ($\mathrm{MRR}$), and two classical univariate rank correlations: Spearman’s $\rho$ \cite{spearman} and Kendall's $\tau$ \cite{kendall}. 

Given a set of $|S|$ cloze sentences and a group of annotators, for each sentence \emph{s}, we denote the list of answers, ranked by their frequency, as $\mathrm{W_\text{s}}=[w_1,w_2,...]$, and the list of model's predictions as $\mathrm{C_\text{s}}=[c_1,c_2,...]$, ranked by their model likelihood. Then, we report $\mathrm{P@k}=\mathbbm{1}[c_i \in \mathrm{W_\text{s}}]$ with $i\in[1,k]$, where $\mathbbm{1}[\cdot]$ is the indicator function. Precision is reported together with its standard deviation, to account for the group-wise disparity in both dimensions (social groups and language): 

\begin{equation}
    \mathrm{\sigma_{gd}} = \sqrt{\frac{\sum_{j=1}^G (\mathrm{P@k}_j - \overline{\mathrm{P@k}})^2}{\mathrm{G}}}
\end{equation}

where $\overline{\mathrm{P@k}}$ is the mean value of all observations, and $\mathrm{G}$ the total number of groups across the dimension fixed each time i.e., $\mathrm{G=4}$ across social groups ({\em MN, FN, MNN, FNN}) and $\mathrm{G=4}$ across languages (EN, ES, DE, FR). We also compute the mean-reciprocal rank ($\mathrm{MRR}$) of the elements of $\mathrm{W_\text{s}}$ with respect to the top-$n$ ($\mathrm{n=5}$) elements of $\mathrm{C_\text{s}}$ ($\mathrm{C^\text{n}_\text{s}}$):

\begin{equation}
    \mathrm{MRR} = \frac{1}{|S|}\sum_{s=1}^{|S|} {\frac{1}{Rank_i^{C^n_s}}}
\end{equation}

Finally, we compute Spearman’s $\rho$ \cite{spearman} and Kendall's $\tau$ \cite{kendall} between $\mathrm{W_\text{s}}$ and $\mathrm{C^\text{5}_\text{s}}$. These metrics are generally more robust to outliers.  

\section{Results}
\label{sec:results}

Following previous work on examining fairness of document classification \cite{huang-etal-2020-multilingual, dixon-etal-2018-mitigate-bias, park-etal-2018-reducing, Garg2019CounterfactualFI}, we focus on group-level performance differences (group disparity). We measure the group disparity as the variance in PLM's performance ($\mathrm{P@k}$) across demographics (gender and native language). Table~\ref{tab:p@1} shows better precision for native speakers in German and French ({\em MN}, {\em FN}) for $\mathrm{P@1}$. In terms of group disparity, male non-natives ({\em MNN}) is the demographic exhibiting the highest disparity across languages in mBERT, while it is female natives ({\em FN}) in XLM-R and male natives ({\em MN}) in mT5. Language-wise, we see the largest group disparity with German in all three models. Here, we see 2.5--4.4 between-group differences, compared to, e.g., 0.3--1.8 between-group differences for English. See Appendix~\ref{sec:appendix} for results with $\mathrm{P@5}$. 

XLM-R consistently exhibits better overall performance on average, but higher between-group and between-language differences in terms of precision ($\sigma_{gd}$). 

\begin{table}[]
\centering
\resizebox{\columnwidth}{!}{
\subfloat{
    \begin{tabular}{l|c|c|c|c|c}
        \multicolumn{6}{c}{\textbf{mBERT}} \\ \cline{2-5}
        {$\mathrm{P@1}$} &
          {\cellcolor{ENC}\textbf{EN}} &
          {\cellcolor{ESC}\textbf{ES}} &
          {\cellcolor{DEC}\textbf{DE}} &
          {\cellcolor{FRC}\textbf{FR}} & \\ \hline
        {\cellcolor{BACK}\textbf{MN}} &
          {\cellcolor{ENC}13.3} &
          {\cellcolor{ESC}12.7} &
          {\cellcolor{DEC}11.3} &
          {\cellcolor{FRC}10.7} & 
          {\cellcolor{BACK}12.0 (1.0)} \\ \hline
        {\cellcolor{BACK}\textbf{FN}} &
          {\cellcolor{ENC}13.3} &
          {\cellcolor{ESC}12.0} &
          {\cellcolor{DEC}15.3} &
          {\cellcolor{FRC}8.0} & 
          {\cellcolor{BACK}12.2 (2.7)} \\ \hline
        {\cellcolor{BACK}\textbf{MNN}} &
          {\cellcolor{ENC_}12.7} &
          {\cellcolor{ESC}12.4} &
          {\cellcolor{DEC}11.4} &
          {\cellcolor{FRC_}3.6} & 
          {\cellcolor{BACK}10.0 \color{red}\textbf{(3.8)}}\\ \hline
        {\cellcolor{BACK}\textbf{FNN}} &
          {\cellcolor{ENC}13.3} &
          {\cellcolor{ESC_}10.0} &
          {\cellcolor{DEC_}5.6} &
          {\cellcolor{FRC_}6.9} & 
          {\cellcolor{BACK}9.0 (3.0) } \\\hline\hline
         &
          {\cellcolor{ENC_}13.2 (0.3)} &
          {\cellcolor{ESC_}11.8 (1.1)} &
          {\cellcolor{DEC_}10.8 {\color{red} \textbf{(3.5)}}} &
          {\cellcolor{FRC_}7.3 (2.5)} & $\overline{\mathrm{P@1}}$($\sigma_{gd}$)
          \end{tabular}
          }
        }
\resizebox{\columnwidth}{!}{
    \subfloat{
    \begin{tabular}{l|c|c|c|c|c}
        \multicolumn{6}{c}{\textbf{XLM-R}} \\ \cline{2-5}
        {$\mathrm{P@1}$} &
          {\cellcolor{ENC}\textbf{EN}} &
          {\cellcolor{ESC}\textbf{ES}} &
          {\cellcolor{DEC}\textbf{DE}} &
          {\cellcolor{FRC}\textbf{FR}} \\ \hline
        {\cellcolor{BACK}\textbf{MN}} &
          {\cellcolor{ENC_}16.7} &
          {\cellcolor{ESC_}13.3} &
          {\cellcolor{DEC}20.7} &
          {\cellcolor{FRC}16.7} & 
          {\cellcolor{BACK}16.9 (2.6)} \\ \hline
        {\cellcolor{BACK}\textbf{FN}} &
          {\cellcolor{ENC_}16.0} &
          {\cellcolor{ESC}15.3} &
          {\cellcolor{DEC}24.0} &
          {\cellcolor{FRC}17.3} & 
          {\cellcolor{BACK}18.2 \color{red}\textbf{(3.5)}}\\ \hline
        {\cellcolor{BACK}\textbf{MNN}} &
          {\cellcolor{ENC_}15.3} &
          {\cellcolor{ESC_}13.5} &
          {\cellcolor{DEC_}15.0} &
          {\cellcolor{FRC_}11.4} & 
          {\cellcolor{BACK}13.8 (1.5)}\\ \hline
        {\cellcolor{BACK}\textbf{FNN}} &
          {\cellcolor{ENC}20.0} &
          {\cellcolor{ESC}14.7} &
          {\cellcolor{DEC_}13.1} &
          {\cellcolor{FRC_}12.7} & 
          {\cellcolor{BACK}15.1 (3.0)} \\\hline\hline
         &
          {\cellcolor{ENC_}17.0 (1.8)} &
          {\cellcolor{ESC_}14.2 (0.8)} &
          {\cellcolor{DEC_}18.2 {\color{red} \textbf{(4.4)}}} &
          {\cellcolor{FRC_}14.5 (2.6)} & $\overline{\mathrm{P@1}}$($\sigma_{gd}$)
        \end{tabular}
        }}
\resizebox{\columnwidth}{!}{
    \subfloat{
    \begin{tabular}{l|c|c|c|c|c}
        \multicolumn{6}{c}{\textbf{mT5}}\\ \cline{2-5}
          {$\mathrm{P@1}$} &
          {\cellcolor{ENC}\textbf{EN}} &
          {\cellcolor{ESC}\textbf{ES}} &
          {\cellcolor{DEC}\textbf{DE}} &
          {\cellcolor{FRC}\textbf{FR}} \\ \hline
        {\cellcolor{BACK}\textbf{MN}} &
          {\cellcolor{ENC_}2.0} &
          {\cellcolor{ESC_}4.7} &
          {\cellcolor{DEC}8.7} &
          {\cellcolor{FRC}5.3} &  
          {\cellcolor{BACK}5.2 \color{red}\textbf{(2.4)}}\\ \hline
        {\cellcolor{BACK}\textbf{FN}} &
          {\cellcolor{ENC}4.0} &
          {\cellcolor{ESC_}3.3} &
          {\cellcolor{DEC}6.7} &
          {\cellcolor{FRC_}3.3} & 
          {\cellcolor{BACK}4.3 (1.4)} \\ \hline
        {\cellcolor{BACK}\textbf{MNN}} &
          {\cellcolor{ENC_}2.0} &
          {\cellcolor{ESC_}4.7} & 
          {\cellcolor{DEC}6.4} &
          {\cellcolor{FRC_}4.3} & 
          {\cellcolor{BACK}4.4 (1.6)} \\\hline
        {\cellcolor{BACK}\textbf{FNN}} &
          {\cellcolor{ENC}3.3} &
          {\cellcolor{ESC}6.7} &
          {\cellcolor{DEC_}1.9} &
          {\cellcolor{FRC}6.2} & 
        {\cellcolor{BACK}4.5 (2.0)} \\\hline\hline  &
          {\cellcolor{ENC_}2.8 (0.9)} &
          {\cellcolor{ESC_}4.8 (1.2)} &
          {\cellcolor{DEC_}5.8 {\color{red} \textbf{(2.5)}}} &
          {\cellcolor{FRC_}4.8 (1.1)} & $\overline{\mathrm{P@1}}$($\sigma_{gd}$)
    \end{tabular} 
    }}
\caption{Results on $\mathrm{P@1}$ score across groups (rows) and languages (columns), average performance in each language ($\overline{\mathrm{P@1}}$) and standard deviation for group disparity ($\sigma_{gd}$). Cells are coloured language-wise. Cells with a darker background are language-wise above the average. Worst group performance in terms of group disparity (highest variance) is highlighted in red.}
\label{tab:p@1}
\end{table}

\begin{figure}[t]
    \centering
    \resizebox{0.95\columnwidth}{!}{
    \includegraphics{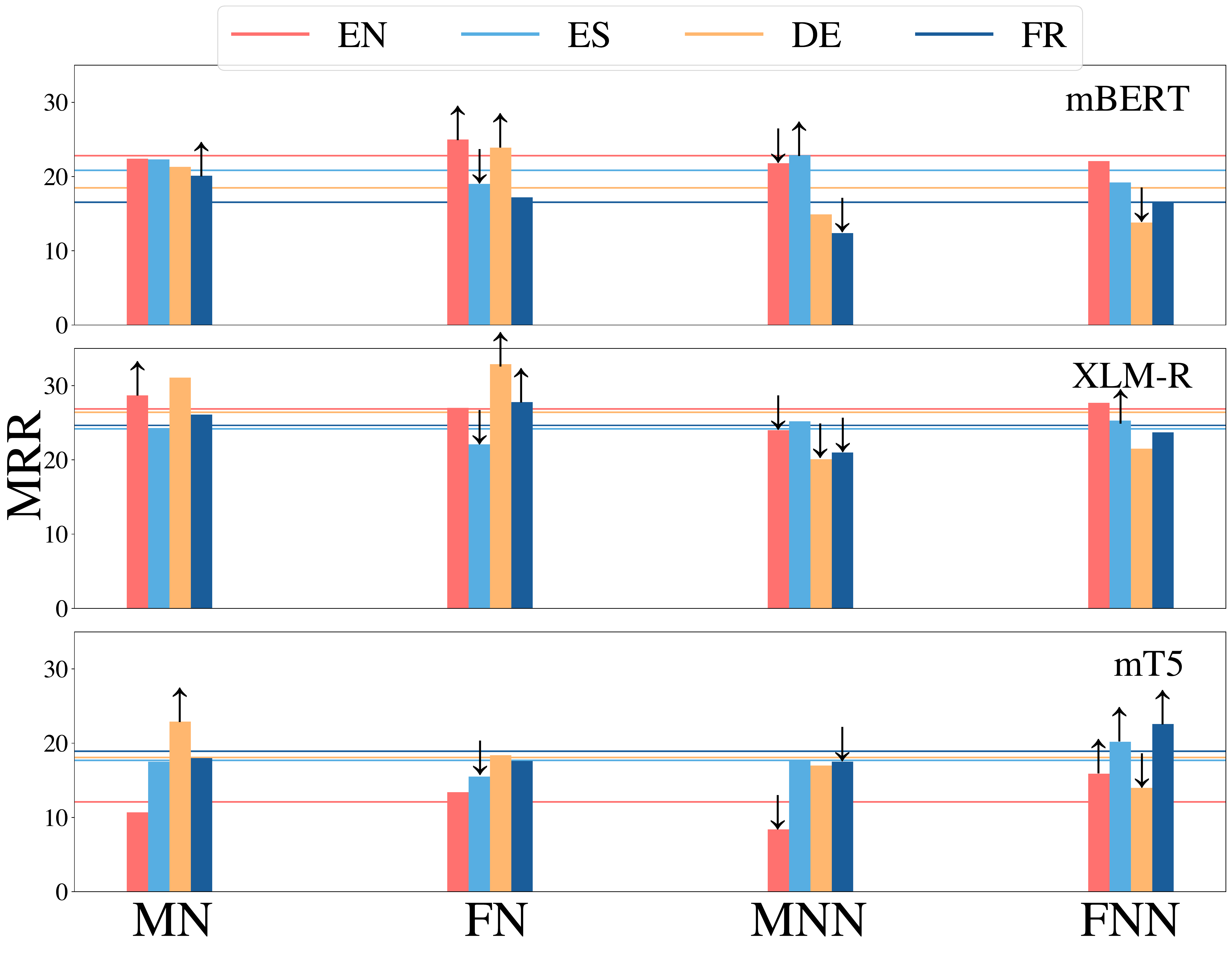}
    }
    \vspace{-2mm}
    \caption{Average $\mathrm{MRR}$ (in percentage) per group in each language. Horizontal lines denote the average per language. Best-off ($\uparrow$) and worst-off ($\downarrow$) subgroups for each language are marked.}
    \label{fig:ranking}
    \vspace{-6mm}
\end{figure}

Figure~\ref{fig:ranking} complements results from Table~\ref{tab:p@1} with $\text{MRR}$ scores. We observe a common trend that the models often underperform on non-native male speakers in all languages except for Spanish: Performance is (always) below the average, and they are the worst-off group ($\downarrow$) in most of the cases. At the same time, predictions with mBERT and XLM-R seem to be biased towards native speakers because answers from {\em MN} and {\em FN} generally rank highest.
Despite none of the models perform equally across groups, XLM-R shows a lower divergence across languages: Between-group differences are more than 50\%~smaller than with mBERT and mT5 when looking at the average $\text{MRR}$ per language.

Table~\ref{tab:spearman} gathers group level Spearman's $\rho$ and average correlation per language. XLM-R predictions are more uniformly correlated across languages compared to mBERT, whose lexical preferences are better aligned in English and Spanish setups, and mT5, whose predictions correlate poorly with human cloze test answers. However, in line with previous results, the model exhibits bias towards male native speakers and {\em MNN} outlines as the worst performing group across languages, with a coefficient always below the average. Looking into the dimension of languages, German is the least aligned with human's answers in all models. Kendall's $\tau$ yields similar results. See Appendix ~\ref{sec:appendix} for details.

\begin{table}[]
\centering
\resizebox{\columnwidth}{!}{
\subfloat{
\begin{tabular}{l|c|c|c|c}
\multicolumn{5}{c}{\textbf{mBERT}} \\ \cline{2-5}
\multicolumn{1}{c|}{$\rho$} &
  {\cellcolor{ENC}\textbf{EN}} &
  {\cellcolor{ESC}\textbf{ES}} &
  {\cellcolor{DEC}\textbf{DE}} &
  {\cellcolor{FRC}\textbf{FR}} \\ \hline
{\cellcolor{BACK}\textbf{MN}} &
  {\cellcolor{ENC}0.33 (p=0.00)} &
  {\cellcolor{ESC}0.23 (p=0.01)} &
  {\cellcolor{DEC_}\color{red}-0.14 (p=0.09)} &
  \cellcolor{FRC_}\color{red}0.10 (p=0.21) \\ \hline
{\cellcolor{BACK}\textbf{FN}} &
  {\cellcolor{ENC_}0.27 (p=0.00)} &
  {\cellcolor{ESC_}\color{red}0.07 (p=0.42)} &
  {\cellcolor{DEC_}\color{red}-0.01 (p=0.89)} &
  \cellcolor{FRC_}\color{red}0.14 (p=0.08) \\ \hline
{\cellcolor{BACK}\textbf{MNN}} &
  {\cellcolor{ENC_}0.30 (p=0.00)} &
  \cellcolor{ESC_}{0.16 (p=0.03)} &
  {\cellcolor{DEC_}\color{red}-0.10 (p=0.23)} &
  \cellcolor{FRC_}\color{red}0.08 (p=0.32) \\ \hline
{\cellcolor{BACK}\textbf{FNN}} &
  {\cellcolor{ENC}0.37 (p=0.00)} &
  {\cellcolor{ESC_}\color{red}0.16 (p=0.06)} &
  {\cellcolor{DEC_}\color{red}0.03 (p=0.69)} &
  \cellcolor{FRC_}\color{red}0.08 (p=0.30) \\ \hline \hline
{\textit{Avg.}} &
  {\cellcolor{ENC_}{0.32 (p=0.00)}} &
  {\cellcolor{ESC_}{0.16 (p=0.00)}} &
  {\cellcolor{DEC_}{\color{red} -0.05 (p=0.21)}} &
  {\cellcolor{FRC_}0.10 (p=0.01)}
\end{tabular}
}}
\resizebox{\columnwidth}{!}{
\subfloat{
\begin{tabular}{l|c|c|c|c}
\multicolumn{5}{c}{\textbf{XLM-R}} \\\cline{2-5}
\multicolumn{1}{c|}{$\rho$} &
  {\cellcolor{ENC}\textbf{EN}} &
  {\cellcolor{ESC}\textbf{ES}} &
  {\cellcolor{DEC}\textbf{DE}} &
  {\cellcolor{FRC}\textbf{FR}} \\ \hline
{\cellcolor{BACK}\textbf{MN}} &
  {\cellcolor{ENC}0.45 (p=0.00)} &
  {\cellcolor{ESC}0.46 (p=0.00)} &
  {\cellcolor{DEC}0.35 (p=0.00)} &
  \cellcolor{FRC}0.48 (p=0.00) \\ \hline
{\cellcolor{BACK}\textbf{FN}} &
  {\cellcolor{ENC_}0.30 (p=0.00)} &
  {\cellcolor{ESC_}0.35 (p=0.00)} &
  {\cellcolor{DEC}0.45 (p=0.00)} &
  \cellcolor{FRC_}0.33 (p=0.00) \\ \hline
{\cellcolor{BACK}\textbf{MNN}} &
  {\cellcolor{ENC_}0.30 (p=0.00)} &
  {\cellcolor{ESC_}0.38 (p=0.00)} &
  {\cellcolor{DEC_}0.22 (p=0.01)} &
  \cellcolor{FRC_}0.32 (p=0.00) \\ \hline
{\cellcolor{BACK}\textbf{FNN}} &
  {\cellcolor{ENC}0.40 (p=0.00)} &
  {\cellcolor{ESC}0.48 (p=0.00)} &
  {\cellcolor{DEC_}\color{red}0.11 (p=0.16)} &
  \cellcolor{FRC_}0.36 (p=0.00) \\ \hline \hline
{\textit{Avg.}} &
  {\cellcolor{ENC_}{0.36 (p=0.00)}} &
  {\cellcolor{ESC_}{0.41 (p=0.00)}} &
  {\cellcolor{DEC_}{0.28 (p=0.00)}} &
  {\cellcolor{FRC_}0.37 (p=0.00)} 
\end{tabular}
}}
\resizebox{\columnwidth}{!}{
\subfloat{
\begin{tabular}{l|c|c|c|c}
  \multicolumn{5}{c}{\textbf{mT5}}\\\cline{2-5}
 \multicolumn{1}{c|}{$\rho$} &
  {\cellcolor{ENC}\textbf{EN}} &
  {\cellcolor{ESC}\textbf{ES}} &
  {\cellcolor{DEC}\textbf{DE}} &
  {\cellcolor{FRC}\textbf{FR}} \\ \hline
{\cellcolor{BACK}\textbf{MN}} &
  {\cellcolor{ENC_}\color{red}0.01 (p=0.89)} &
  {\cellcolor{ESC_}\color{red}0.14 (p=0.08)} &
  {\cellcolor{DEC_}\color{red}0.14 (p=0.08)} &
  \cellcolor{FRC}0.25 (p=0.00) \\ \hline
{\cellcolor{BACK}\textbf{FN}} &
  {\cellcolor{ENC_}\color{red}-0.12 (p=0.13)} &
  {\cellcolor{ESC_}\color{red}0.13 (p=0.12)} &
  {\cellcolor{DEC_}\color{red}0.00 (p=0.99)} &
  \cellcolor{FRC_}\color{red}0.14 (p=0.08) \\ \hline
{\cellcolor{BACK}\textbf{MNN}} &
  {\cellcolor{ENC_}\color{red}-0.10 (p=0.22)} &
  {\cellcolor{ESC_}\color{red}0.12 (p=0.11)} &
  {\cellcolor{DEC_}\color{red}0.03 (p=0.74)} &
  \cellcolor{FRC_}\color{red}0.11 (p=0.18) \\ \hline
{\cellcolor{BACK}\textbf{FNN}} &
  {\cellcolor{ENC_}\color{red}-0.07 (p=0.41)} &
  {\cellcolor{ESC}0.28 (p=0.00)} &
  {\cellcolor{DEC_}\color{red}0.04 (p=0.58)} &
  \cellcolor{FRC_}\color{red}0.11 (p=0.16) \\ \hline \hline
{\textit{Avg.}} &
  {\cellcolor{ENC_}{\color{red}-0.07 (p=0.07)}} &
  {\cellcolor{ESC_}{0.17 (p=0.00)}} &
  {\cellcolor{DEC_}{\color{red}0.05 (p=0.23)}} &
  {\cellcolor{FRC_}0.15 (p=0.00)}
\end{tabular}}}
\caption{Correlation between groups of annotators ({\em MN, FN, MNN, FNN}) and models' predictions, classified by language. The degree of correlation is measured with Spearman's $\rho$ coefficient ($\mathrm{\rho\in[-1,1]}$). Cells are coloured language-wise. Cells with a darker background show a stronger correlation compared to the average in each language. Samples highlighted in red fail to reject the null hypothesis, meaning that their difference is not statistically significant ($\mathrm{p>0.05}$). }
\label{tab:spearman}
\end{table}

It is worth mentioning that our study does not aim to compare models' performance, but rather to motivate a discussion about the between-group and between-language differences within each model. The general low precision of mT5 outputs compared to human answers is likely due to the nature of the task itself. Because mT5 was trained with a span-mask denoising objective, it tends to complete the masked-out span with more than one token. When constraining generation to output one token, we are conditioning its default behaviour. Better correlation could be achieved by fine-tuning the model on completing cloze tests.

(Dis)agreement amongst annotators on the same language gives a measure of the difficulty of the task. French and German present a higher variability in the responses (with a vocabulary of 442 and 443 words respectively), compared to English (374 words), and Spanish (427 words), which reflects in a lower correlation with models' predictions.


\section{Related Work}

Multilingual PLMs have been analyzed in many ways: Researchers have, for example, looked at performance differences across languages \cite{singh-etal-2019-bert, wu-dredze-2020-languages}, looked at their organization of language types \cite{rama-etal-2020-probing}, used similarity analysis to probe their representations \cite{kudugunta-etal-2019-investigating}, and investigated how learned self-attention in the Transformer blocks affects different languages \cite{ravishankar-etal-2021-attention}. 

Previous work on fairness of multilingual models has, to the best of our knowledge, focused exclusively on task-specific models, rather than PLMs: \citet{huang-etal-2020-multilingual} evaluate the fairness of multilingual hate speech detection models, and several researchers have explored gender bias in multilingual models \cite{zhao-etal-2020-gender,gonzalez-etal-2020-type}. \citet{dayanik-pado-2021-disentangling} consider the effects of adversarial debiasing in multilingual models. 

Cloze tests were previously used in \citet{zhang-etal-2021-sociolectal} to evaluate the fairness of English (monolingual) language models. In psycholinguistics, cloze tests have been performed with different age groups \cite{hintz-etal-2020-behavioral} and native language \cite{stringer-iverson-2020-nonnative}, but these datasets have, to the best of our knowledge, not been used to evaluate language models.

\section{Conclusion}
\label{sec:conclusions}

In this paper, we present MozArt, a new multilingual dataset of parallel cloze examples with annotations from balanced demographics. This dataset is, to the best of our knowledge, the first to enable apples-to-apples comparison of group disparity of multilingual PLMs across languages. The dataset includes several demographic attributes, but we present preliminary experiments with gender and native language. We show that mBERT, XLM-R and mT5 are not equally fair across languages. For example, group disparities are much higher for German (and French) than for English and Spanish. This shows the importance of evaluating fairness across languages instead of stipulating from results for a single language.  
We further show that cloze test answers of female native speakers tend to rank highest in both predictive PLMs. We followed best practices for mitigating the dangers of crowdsourcing \cite{karpinska2021perils, suzanne-kleijn} (see \S2) and hope MozArt will be widely adopted and, over time, generate more results for other languages, PLMs and demographic attributes. 

\section{Limitations}
\label{sec:limitations}

As described in the paper, MozArt builds on top of another dataset, which is only available in four languages. The original dataset with its manual word alignments provided a unique opportunity to build MozArt in a way in which we could account for context, across languages. This of course limits our work to the languages provided. We acknowledge how multilingual studies of Indo-European languages may not generalize to languages outside this language famility, and hope we or others will be able to contribute resources for a more diverse set of languages in the future.

\section*{Ethics Statement}
\label{sec:ethics}

The dataset released contains publicly available content from the proceedings of the European Parliament. Our work is based on sensitive information provided by the participants that took on our study in Prolific. The protected attributes collected are self-reported on a voluntary basis, and participants gave their consent to share them. In addition to the specific attributes analyzed in our study, which served as prescreening filters, Prolific also provides baseline data for all studies with the consent of participants to share it with researchers. For these base attributes, there might be gaps in the data because it is optional for participants to provide this information. These attributes are filled as {\em null} in the dataset. We performed a pilot study to determine the amount of time a task would take on average. The participants were paid based on time worked, and were given the option to opt out at any time of the study. Participants who revoked consent at any stage are not included in our study nor in the data released.

\bibliography{anthology}
\bibstyle{acl_natbib}

\appendix

\section{Additional results}
\label{sec:appendix}

In this section, we provide additional analysis results of the PLM's performance on MozArt. We report precision at 5 (P@5), which corresponds to the number of relevant answers amongst the top 5 candidates. It provides a more flexible metric for measuring model alignments with open-ended text answers, but fails to take into account the exact position within the top-k. Considering the top-5, the bias towards native speakers is diminished especially in English and Spanish, despite being {\em MNN} and {\em FNN} the worst groups --in terms of group disparity-- in mBERT and XLM-R respectively. At the same time, the group disparities are exacerbated as shown in Table~\ref{tab:p@5}.

\begin{table}[!h]
\centering
\resizebox{\columnwidth}{!}{
\subfloat{
\begin{tabular}{l|c|c|c|c|c}
\multicolumn{6}{c}{\textbf{mBERT}} \\ \cline{2-5}
{$\mathrm{P@5}$} &
  {\cellcolor{ENC}\textbf{EN}} &
  {\cellcolor{ESC}\textbf{ES}} &
  {\cellcolor{DEC}\textbf{DE}} &
  {\cellcolor{FRC}\textbf{FR}} & \\ \hline
{\cellcolor{BACK}\textbf{MN}} &
  {\cellcolor{ENC_}30.7} &
  {\cellcolor{ESC}26.7} &
  {\cellcolor{DEC}22.0} &
  {\cellcolor{FRC}24.0} & 
  {\cellcolor{BACK}25.9 (3.3)} \\ \hline
{\cellcolor{BACK}\textbf{FN}} &
  {\cellcolor{ENC_}32.0} &
  {\cellcolor{ESC_}18.7} &
  {\cellcolor{DEC}24.7} &
  {\cellcolor{FRC}22.0} & 
  {\cellcolor{BACK}24.4 (4.9)} \\ \hline
{\cellcolor{BACK}\textbf{MNN}} &
  {\cellcolor{ENC}34.0} &
  {\cellcolor{ESC}25.9} &
  {\cellcolor{DEC_}12.1} &
  {\cellcolor{FRC_}15.0} & 
  {\cellcolor{BACK}21.8 \color{red}\textbf{(8.7)}}\\ \hline
{\cellcolor{BACK}\textbf{FNN}} &
  {\cellcolor{ENC}32.7} &
  {\cellcolor{ESC}25.3} &
  {\cellcolor{DEC_}16.3} &
  {\cellcolor{FRC_}16.3} & 
  {\cellcolor{BACK}22.7 (6.9) } \\\hline\hline
 &
  {\cellcolor{ENC_}32.3 (1.2)} &
  {\cellcolor{ESC_}24.2 (3.1)} &
  {\cellcolor{DEC_}18.8 {\color{red} \textbf{(4.9)}}} &
  {\cellcolor{FRC_}19.3 (3.8)} & {$\overline{\mathrm{P@5}}$($\sigma_{gd}$)}
\end{tabular}}}
\resizebox{\columnwidth}{!}{
\subfloat{
\begin{tabular}{l|c|c|c|c|c}
\multicolumn{6}{c}{\textbf{XLM-R}} \\ \cline{2-5}
{$\mathrm{P@5}$} &
  {\cellcolor{ENC}\textbf{EN}} &
  {\cellcolor{ESC}\textbf{ES}} &
  {\cellcolor{DEC}\textbf{DE}} &
  {\cellcolor{FRC}\textbf{FR}} \\ \hline
{\cellcolor{BACK}\textbf{MN}} &
  {\cellcolor{ENC}39.3} &
  {\cellcolor{ESC}30.7} &
  {\cellcolor{DEC}34.7} &
  {\cellcolor{FRC}32.7} & 
  {\cellcolor{BACK}34.4 (3.2)} \\ \hline
{\cellcolor{BACK}\textbf{FN}} &
  {\cellcolor{ENC}30.7} &
  {\cellcolor{ESC_}25.3} &
  {\cellcolor{DEC}38.0} &
  {\cellcolor{FRC}35.3} & 
  {\cellcolor{BACK}32.3 (4.8)}\\ \hline
{\cellcolor{BACK}\textbf{MNN}} &
  {\cellcolor{ENC_}30.7} &
  {\cellcolor{ESC_}29.4} &
  {\cellcolor{DEC_}22.1} &
  {\cellcolor{FRC_}25.4} & 
  {\cellcolor{BACK}26.9 (3.4)}\\ \hline
{\cellcolor{BACK}\textbf{FNN}} &
  {\cellcolor{ENC}36.7} &
  {\cellcolor{ESC}34.0} &
  {\cellcolor{DEC_}19.4} &
  {\cellcolor{FRC_}26.9} & 
  {\cellcolor{BACK}29.3 {\color{red}\textbf{(6.7)}}} \\\hline\hline
 &
  {\cellcolor{ENC_}34.3 (3.8)} &
  {\cellcolor{ESC_}29.8 (3.1)} &
  {\cellcolor{DEC_}28.5 {\color{red} \textbf{(7.9)}}} &
  {\cellcolor{FRC_}30.3 (4.1)} & {$\overline{\mathrm{P@5}}$($\sigma_{gd}$)}
\end{tabular}}}
\resizebox{\columnwidth}{!}{
\subfloat{
\begin{tabular}{l|c|c|c|c|c}
\multicolumn{6}{c}{\textbf{mT5}}\\ \cline{2-5}
  {$\mathrm{P@5}$} &
  {\cellcolor{ENC}\textbf{EN}} &
  {\cellcolor{ESC}\textbf{ES}} &
  {\cellcolor{DEC}\textbf{DE}} &
  {\cellcolor{FRC}\textbf{FR}} \\ \hline
{\cellcolor{BACK}\textbf{MN}} &
  {\cellcolor{ENC_}10.0} &
  {\cellcolor{ESC}12.7} &
  {\cellcolor{DEC}16.0} &
  {\cellcolor{FRC_}11.3} &  
  {\cellcolor{BACK}12.5 (2.2)}\\ \hline
{\cellcolor{BACK}\textbf{FN}} &
  {\cellcolor{ENC}11.3} &
  {\cellcolor{ESC_}10.0} &
  {\cellcolor{DEC}16.7} &
  {\cellcolor{FRC}18.0} & 
  {\cellcolor{BACK}14.0 \color{red}\textbf{(3.4)}} \\ \hline
{\cellcolor{BACK}\textbf{MNN}} &
  {\cellcolor{ENC_}6.0} &
  {\cellcolor{ESC_}11.8} & 
  {\cellcolor{DEC_}9.3} &
  {\cellcolor{FRC_}10.7} & 
  {\cellcolor{BACK}9.5 (2.2)} \\\hline
{\cellcolor{BACK}\textbf{FNN}} &
  {\cellcolor{ENC}13.3} &
  {\cellcolor{ESC}16.0} &
  {\cellcolor{DEC_}8.7} &
  {\cellcolor{FRC}15.0} & 
  {\cellcolor{BACK}13.3 (2.8)} \\\hline\hline
&
  {\cellcolor{ENC_}10.2 (2.7)} &
  {\cellcolor{ESC_}12.6 (2.2)} &
  {\cellcolor{DEC_}12.7 {\color{red} \textbf{(3.7)}}} &
  {\cellcolor{FRC_}13.8 (3.0)} & {$\overline{\mathrm{P@5}}$($\sigma_{gd}$)} 
\end{tabular}}}
\caption{Results on $\mathrm{P@5}$ score across groups and languages, average performance in each language ($\overline{\mathrm{P@5}}$) and standard deviation for group disparity ($\sigma_{gd}$). Cells are coloured language-wise. Cells with a darker background are language-wise above the average. Worst group performance in terms of group disparity (highest variance) is highlighted in red.}
\label{tab:p@5}
\end{table}

Table~\ref{tab:kendall} complements results on correlation of the alignment of group responses. It shows Kendall's $\tau$ coefficient. Conclusions remain almost the same as studied with Spearman's coefficient, albeit non-native subgroups in Spanish are more correlated in mBERT.

\begin{table}[]
\centering
\resizebox{\columnwidth}{!}{
\subfloat{
\begin{tabular}{l|c|c|c|c}
\multicolumn{5}{c}{\textbf{mBERT}} \\ \cline{2-5}
 \multicolumn{1}{c|}{$\tau$} &
 {\cellcolor{ENC}\textbf{EN}} &
 {\cellcolor{ESC}\textbf{ES}} &
 {\cellcolor{DEC}\textbf{DE}} &
  {\cellcolor{FRC}\textbf{FR}} \\ \hline
{\cellcolor{BACK}\textbf{MN}} &
 {\cellcolor{ENC}0.27 (p=0.00)} &
 {\cellcolor{ESC}0.19 (p=0.00)} &
 {\cellcolor{DEC_}\color{red}-0.09 (p=0.15)} &
 \cellcolor{FRC_}\color{red}0.09 (p=0.16) \\ \hline
{\cellcolor{BACK}\textbf{FN}} &
 {\cellcolor{ENC_}0.23 (p=0.00)} &
 {\cellcolor{ESC_}\color{red}0.07 (p=0.24)} &
 {\cellcolor{DEC_}\color{red}0.01 (p=0.89)} &
  {\cellcolor{FRC}0.13 (p=0.04)} \\ \hline
{\cellcolor{BACK}\textbf{MNN}} &
 {\cellcolor{ENC_}0.25 (p=0.00)} &
 {\cellcolor{ESC}0.15 (p=0.01)} &
 {\cellcolor{DEC_}\color{red}-0.06 (p=0.32)} &
  \cellcolor{FRC_}\color{red}0.07 (p=0.28) \\ \hline
{\cellcolor{BACK}\textbf{FNN}} &
 {\cellcolor{ENC}0.29 (p=0.00)} &
 {\cellcolor{ESC_}0.14 (p=0.01)} &
 {\cellcolor{DEC_}\color{red}0.03 (p=0.57)} &
  \cellcolor{FRC_}\color{red}0.06 (p=0.27) \\ \hline \hline
 \multicolumn{1}{c|}{\textit{Avg.}} &
 {\cellcolor{ENC_}{ 0.26 (p=0.00)}} &
 {\cellcolor{ESC_}{ 0.14 (p=0.00)}} &
 {\cellcolor{DEC_}{\color{red} -0.03 (p=0.41)}} &
  {\cellcolor{FRC_} 0.09 (p=0.01)}

\end{tabular}}}
\resizebox{\columnwidth}{!}{
\subfloat{
\begin{tabular}{l|c|c|c|c}
\multicolumn{5}{c}{\textbf{XLM-R}} \\ \cline{2-5}
 \multicolumn{1}{c|}{$\tau$} &
 {\cellcolor{ENC}\textbf{EN}} &
 {\cellcolor{ESC}\textbf{ES}} &
 {\cellcolor{DEC}\textbf{DE}} &
  {\cellcolor{FRC}\textbf{FR}} \\ \hline
{\cellcolor{BACK}\textbf{MN}} &
 {\cellcolor{ENC}0.40 (p=0.00)} &
 {\cellcolor{ESC}0.43 (p=0.00)} &
 {\cellcolor{DEC}0.32 (p=0.00)} &
  \cellcolor{FRC}0.45 (p=0.00) \\ \hline
{\cellcolor{BACK}\textbf{FN}} &
 {\cellcolor{ENC_}0.26 (p=0.00)} &
 {\cellcolor{ESC_}0.33 (p=0.00)} &
 {\cellcolor{DEC}0.43 (p=0.00)} &
  \cellcolor{FRC_}0.31 (p=0.00) \\ \hline
{\cellcolor{BACK}\textbf{MNN}} &
 {\cellcolor{ENC_}0.26 (p=0.00)} &
 {\cellcolor{ESC_}0.35 (p=0.00)} &
 {\cellcolor{DEC_}0.20 (p=0.01)} &
  \cellcolor{FRC_}0.29 (p=0.00) \\ \hline
{\cellcolor{BACK}\textbf{FNN}} &
 {\cellcolor{ENC}0.35 (p=0.00)} &
 {\cellcolor{ESC}0.45 (p=0.00)} &
 {\cellcolor{DEC_}\color{red}0.10 (p=0.15)} &
  \cellcolor{FRC_}0.34 (p=0.00) \\ \hline \hline
\multicolumn{1}{c|}{\textit{Avg.}} &
 {{ 0.32 (p=0.00)}} &
 {{ 0.39 (p=0.00)}} &
 {{ 0.25 (p=0.00)}} &
  {\cellcolor{FRC_} 0.34 (p=0.00)} 
\end{tabular}}}
\resizebox{\columnwidth}{!}{
\subfloat{
\begin{tabular}{l|c|c|c|c}

  \multicolumn{5}{c}{\textbf{mT5}}\\ \cline{2-5}
  \multicolumn{1}{c|}{$\tau$} &
 {\cellcolor{ENC}\textbf{EN}} &
 {\cellcolor{ESC}\textbf{ES}} &
 {\cellcolor{DEC}\textbf{DE}} &
  {\cellcolor{FRC}\textbf{FR}} \\ \hline
{\cellcolor{BACK}\textbf{MN}} &
 {\cellcolor{ENC_}\color{red}0.02 (p=0.79)} &
 {\cellcolor{ESC_}\color{red}0.13 (p=0.06)} &
 {\cellcolor{DEC_}\color{red}0.13 (p=0.06)} &
  \cellcolor{FRC}0.21 (p=0.00) \\ \hline
{\cellcolor{BACK}\textbf{FN}} &
 {\cellcolor{ENC_}\color{red}-0.09 (p=0.16)} &
 {\cellcolor{ESC_}\color{red}0.11 (p=0.11)} &
 {\cellcolor{DEC_}\color{red}0.00 (p=0.98)} &
  \cellcolor{FRC_}\color{red}0.12 (p=0.08) \\ \hline
{\cellcolor{BACK}\textbf{MNN}} &
 {\cellcolor{ENC_}\color{red}-0.08 (p=0.21)} &
 {\cellcolor{ESC_}\color{red}0.10 (p=0.10)} &
 {\cellcolor{DEC_}\color{red}0.03 (p=0.69)} &
  \cellcolor{FRC_}\color{red}0.10 (p=0.17) \\ \hline
{\cellcolor{BACK}\textbf{FNN}} &
 {\cellcolor{ENC_}\color{red}-0.04 (p=0.51)} &
 {\cellcolor{ESC}0.25 (p=0.00)} &
 {\cellcolor{DEC_}\color{red}0.03 (p=0.61)} &
 \cellcolor{FRC_} \color{red}0.10 (p=0.15) \\ \hline \hline
\multicolumn{1}{c|}{\textit{Avg.}} &
 {\cellcolor{ENC_}{\color{red}-0.07 (p=0.07)}} &
 {\cellcolor{ESC_}{0.15 (p=0.00)}} &
 {\cellcolor{DEC_}{\color{red}0.05 (p=0.18)}} &
  \cellcolor{FRC_}{0.13 (p=0.00)}
\end{tabular}}}
\caption{Correlation between groups of annotators ({\em MN, FN, MNN, FNN}) and models' predictions, classified by language. The degree of correlation is measured with Kendall's $\tau$ coefficient ($\tau\in[-1,1]$). Cells are coloured language-wise. Cells with a darker background show a stronger correlation compared to the average in each language. Samples highlighted in red fail to reject the null hypothesis, meaning that their difference is not statistically significant ($\mathrm{p>0.05}$). }
\label{tab:kendall}
\end{table}

\section{t-SNE}
\label{sec:appendix_tsne}

\begin{figure}[t]
    \centering
    \resizebox{0.95\columnwidth}{!}{
    \includegraphics{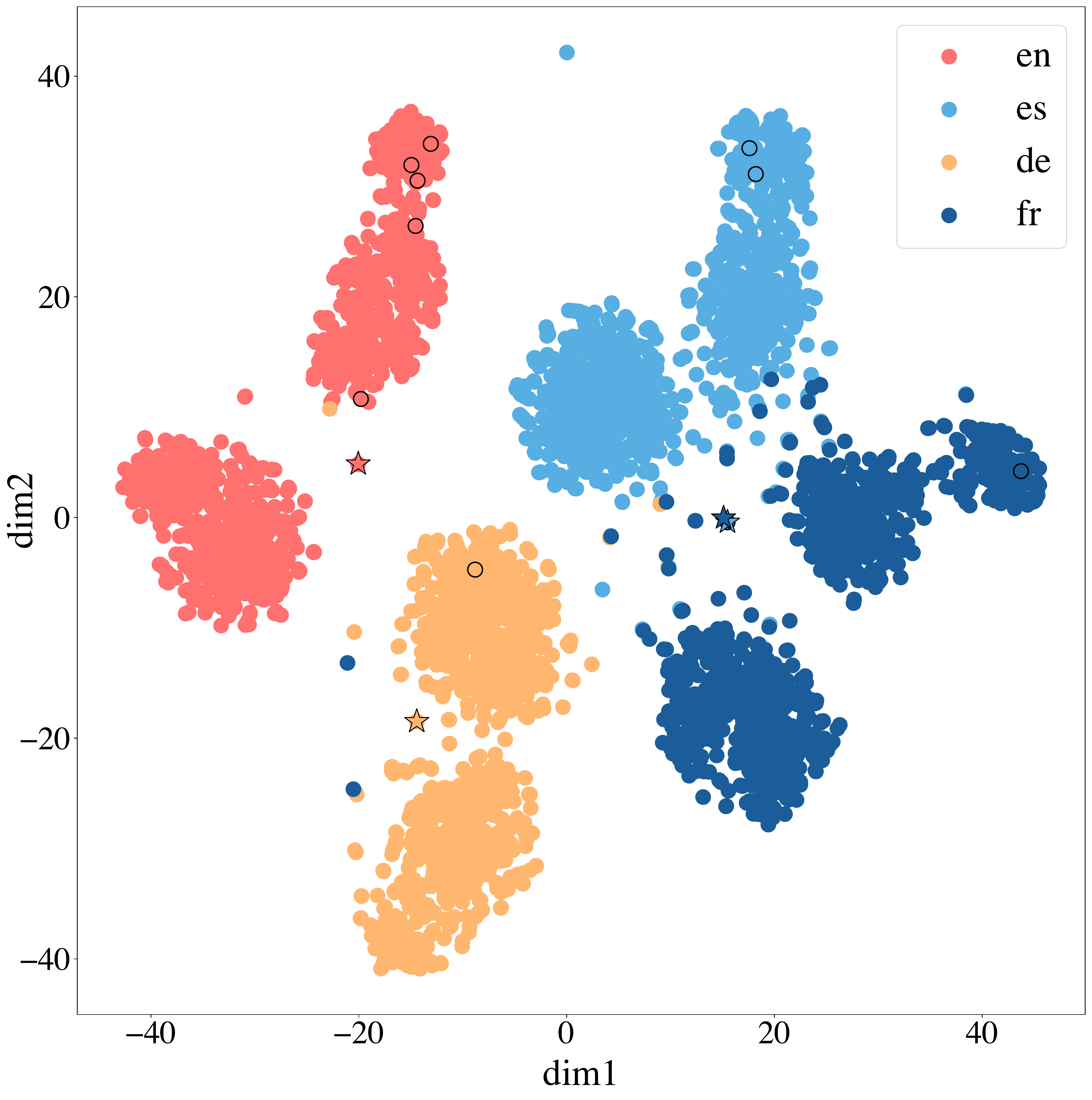}
    }
    \vspace{-2mm}
    \caption{t-SNE representation from the last layer of mBERT for the top-1000 predictions for the parallel sentences in the list above (``We want to [MASK] innovation .'' in English). Highest scored prediction is starred; annotator's answers are denoted by a dot with black edge. Legend shows language-color mapping.}
    \label{fig:tsne}
    \vspace{-6mm}
\end{figure}

\begin{figure}[t]
    \centering
    \resizebox{0.95\columnwidth}{!}{
    \includegraphics{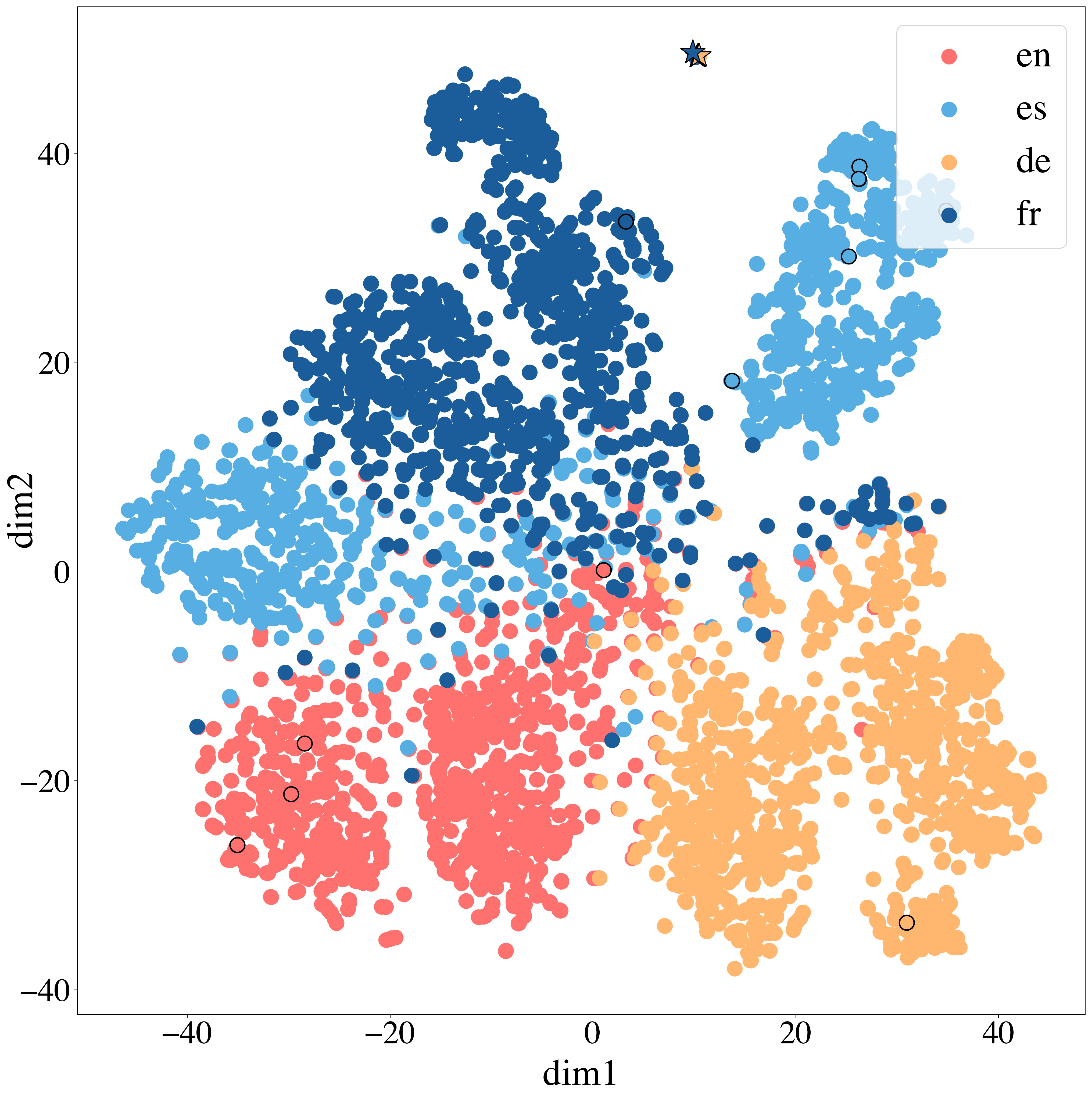}
    }
    \vspace{-2mm}
    \caption{t-SNE representation from the last layer of XLM-R for the top-1000 predictions for the parallel sentences in the list above (``We want to [MASK] innovation .'' in English). Highest scored prediction is starred; annotator's answers are denoted by a dot with black edge. Legend shows language-color mapping.}
    \label{fig:tsnexlmr}
    \vspace{-6mm}
\end{figure}

To give a brief overview of the semantic multilinguality encoded in the pretrained models, we run several representations with t-SNE. Figure~\ref{fig:tsne} and Figure~\ref{fig:tsnexlmr}  represent the top-1000 predictions in a t-SNE plot for mBERT and XLM-R respectively. The same sentence is queried to the model in four languages and, accordingly, to annotators:\\[1pt]

{\resizebox{0.95\columnwidth}{!}{
\begin{tabular}{ll}
    en&We want to [MASK] innovation .\\
    es&Queremos [MASK] la innovación .\\
    de&Wir wollen zur Innovation [MASK] .\\
    fr&Nous voulons [MASK] l'innovation .
\end{tabular} 
}}

\bigskip
\noindent Highest scored predictions are highlighted with a ($\star$). Annotator's answers that fell into the top-1000 predictions are denoted with a black edge. In line with results in \cite{choenni2020does}, we observe that languages are mostly projected in separate sub-spaces instead of yielding a neutral representation, even though they share a common space (vocabulary). 

\begin{figure}[t]
    \centering
    \resizebox{0.95\columnwidth}{!}{
    \includegraphics{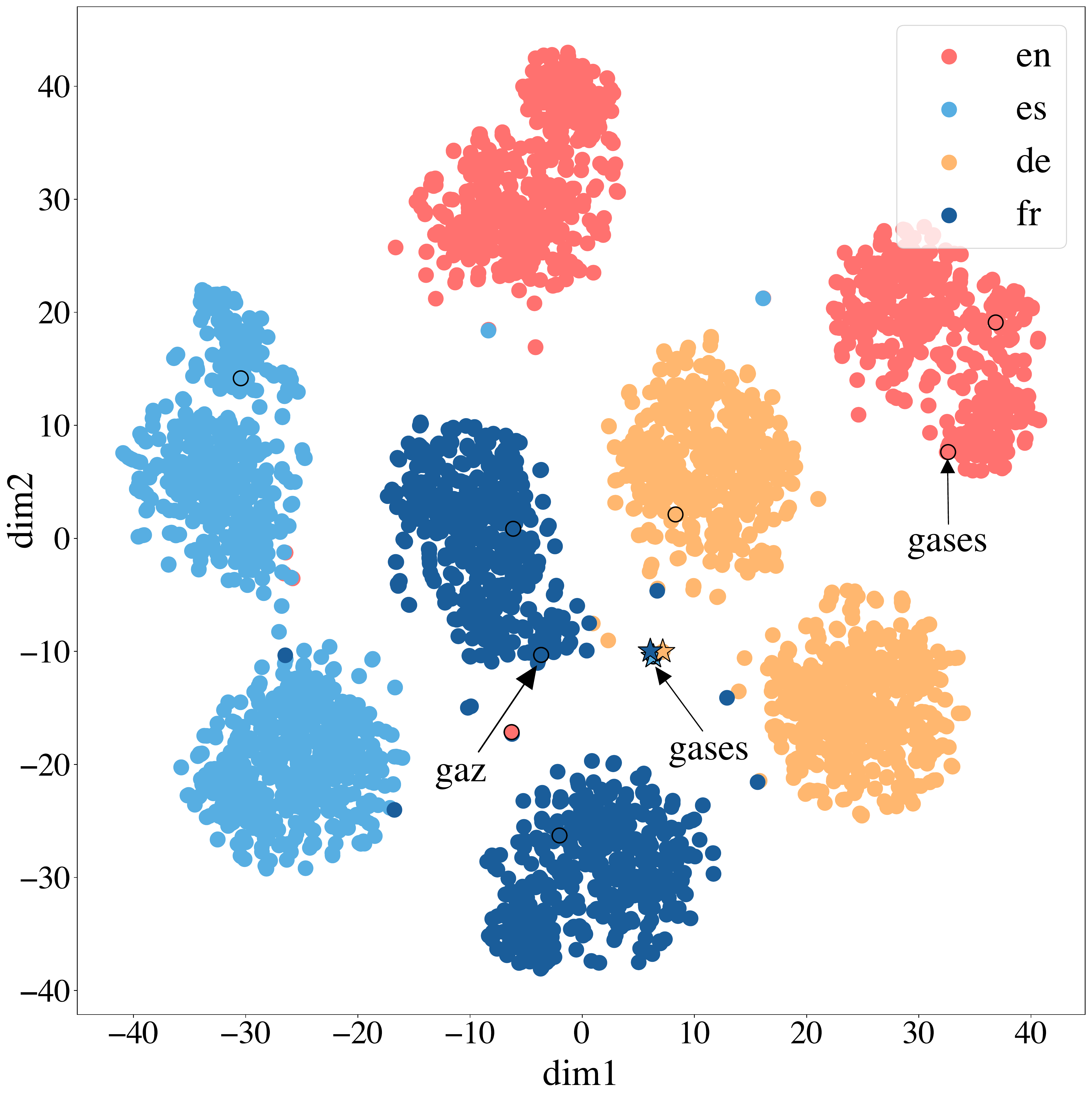}
    }
    \vspace{-2mm}
    \caption{t-SNE representation from the last layer of mBERT for the top-1000 predictions for the parallel sentences in the list above (``[MASK] that deplete the ozone layer'' in English). The word ``gases'' is pointed out in each language (en: gases, es: gases, fr: gaz), as it was a recurrent answer from different annotators. Highest scored prediction in each language is starred; annotator's answers are denoted by a dot with black edge. Legend shows language-color mapping.}
    \label{fig:tsne2}
    \vspace{-6mm}
\end{figure}

\begin{figure}[t]
    \centering
    \resizebox{0.95\columnwidth}{!}{
    \includegraphics{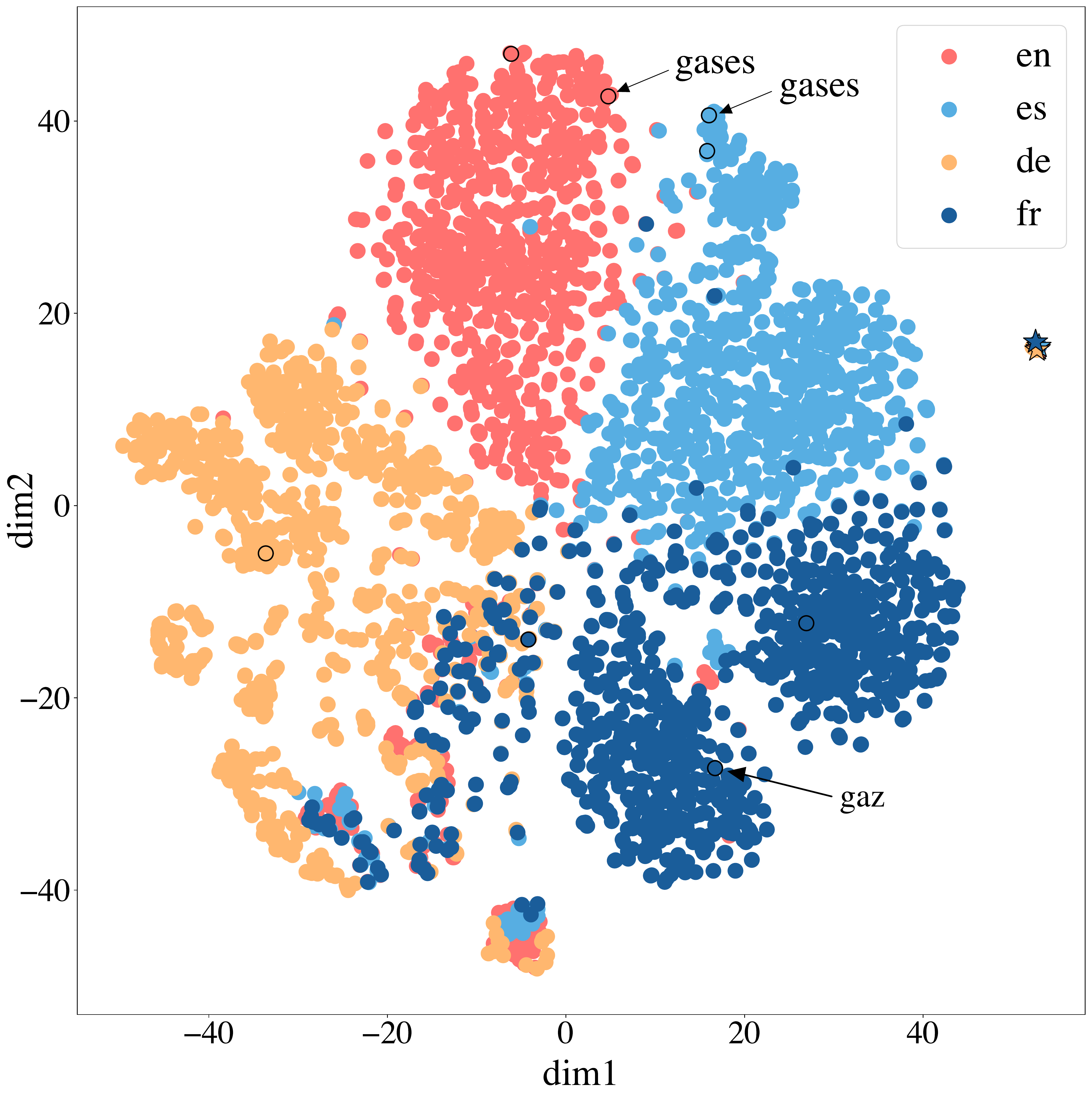}
    }
    \vspace{-2mm}
    \caption{t-SNE representation from the last layer of XLM-R for the top-1000 predictions for the parallel sentences in the list above (``[MASK] that deplete the ozone layer'' in English). The word ``gases'' is pointed out in each language (en: gases, es: gases, fr: gaz), as it was a recurrent answer from different annotators. Highest scored prediction in each language is starred; annotator's answers are denoted by a dot with black edge. Legend shows language-color mapping.}
    \label{fig:tsne2xlmr}
    \vspace{-6mm}
\end{figure}

Similarly, \citet{singh-etal-2019-bert} shown a trend towards dissimilarity between representations for semantically similar inputs in different languages, in deeper layers of an uncased mBERT. Serve Figure~\ref{fig:tsne2} as an example, where the same word ``gases'' was answered in different languages but is represented in different subspaces. Figure~\ref{fig:tsne2xlmr} shows a similar behaviour in XLM-R. The sentences queried are:\\[1pt]

{\resizebox{0.97\columnwidth}{!}{
\begin{tabular}{ll}
    en&[MASK] that deplete the ozone layer\\
    es&[MASK] que agotan la capa de ozono\\
    de&[MASK], die zum Abbau der\\& Ozonschicht führen\\
    fr&[MASK] appauvrissant la couche d'ozone
\end{tabular} 
}}

\end{document}